# The Syntax of Wh-agreement in Yemeni Ibbi Arabic

Ashraf Naji, Ibb University
Mohammed Q. Shormani, Ibb University
shormani@ibbuniv.edu.ye/https://orcid.org/0000-0002-0138-4793

## Abstract

This article tackles an important phenomenon in the syntax of Yemeni Ibbi Arabic (YIA), viz., wh-agreement, a phenomenon common to several languages including Greek, Indonesian, Lubukusu, Irish, etc. In YIA, wh-agreement manifests itself via agreement inflections on the Wh-Op, C, T/V, v. To account for this phenomenon, we propose an Agree across phases (AAP) approach anchored in the mechanism of Feature Inheritance (FI) in which Agree as MATCHING is a bit separated from feature valuation (FV). AM concerns Cs/vs, but FV Ts/Vs. Analyzing the agreement patterns observed between Wh-Op(erators), functional heads (precisely C, (T), v), and verbal complexes, we argue that the suffixes -eh, -uh, -nen, -um, having undergone grammaticalization process from Stannard Arabic (SA) third person pronouns, function as morphological marking of wh-agreement. Findings indicate that YIA data offer a unique empirical contribution to generative syntax, specifically concerning wh-agreement in this dialect operating via MATCHING mechanism. Our proposal straightforwardly accounts for wh-agreement cross-linguistically. This study provides further evidence that incorporating under-investigated typology provides further support for the universality of Universal Grammar (UG) by revealing how specific I-language operations reflect deeper, invariant principles of human language architecture. It concludes that the wh-agreement mechanism in YIA is more morphosyntactically robust than in languages such as Greek, Indonesian, Palauan, and Irish, providing compelling evidence for AAP as a UG approach to long-distance dependencies.

## 1. Introduction

In generative enterprise, the syntax of wh-agreement has received much less attention compared to other phenomena. This is perhaps due to the fact that wh-agreement is found in few languages (see e.g. Jeoung 2020, for Indonesian, Diercks 2013, for Lubukusu, Chung 1994, for Chamorro, Lochbihler and Mathieu 2013, for Ojibwe). Yemeni Ibbi Arabic (YIA) manifests a wh-agreement phenomenon resembling the one found in the languages just cited. YIA is a dialect of Arabic spoken in Yemen. This language variety has several subdialects such as San'ani, Adeni, Ibbi, Thamari, etc. In this study, we focus on Ibbi dialect, which is spoken in Ibb Governorate, Yemen. Although Yemeni Arabic, in general, is not investigated or paid attention to by linguists like many other modern varieties of Arabic, there are few studies addressing some aspects of the grammar such as syntax, phonology, morphology, etc. in some dialects of Yemeni Arabic such as San'ani, Adeni, Hadrami. However, Ibbi Arabic dialect is a very under-investigated language variety (see also Shormani 2019). Ibbi dialect is a phonologically, morphologically, syntactically rich language variety. It involves linguistic "secrets" that are of much importance to the field, and that should be investigated by linguists interested in deeply probing its linguistically rich typology.

One such syntactic phenomenon is wh-agreement as examples (1 & 2) show. However, it seems that this wh-agreement is linked to many other phenomena. Examples in (1 & 2) raise several questions including: i) What is the suffix *-e(h)/-uh* on both the Wh-Op(erator)s and verbs?, ii) If it marks agreement, What type of agreement is it in (1) and (2)?, iii) What is *li-* in (2a)?, and iv) How can we account for the agreement in (2b) and longer structures.

(1) a. man-eh jāʔ-ah ?
who-3FS came-3FS
'Who came?'
b. man-eh qabal-eh ʕali?
who-3FS meet-3FS Ali
'Who did Ali meet?'

(2) a. mu-(h)u li-ʔaštri-uh ʕali?
what-3MS that-bought-it-3MS Ali
'What did Ali buy?'
b. ʔayyan kitab qāl ʕali ʔinn-uh ʔaštri-uh l-ha?
which book said.3MS Ali that-3MS bought-3MS for-3FS
'Which book did Ali that that he bought for her?'

To introduce the study problem, thus, let us touch on these questions. As for (i) and (ii), it seems that the suffix *-e(h)/-u(h)* is an agreement inflection, marking wh-agreement. Regarding (iii), *li-* is likely a reduced relative pronoun (RRP). RRP is a well-known notion in colloquial Arabic (see e.g. Ouhalla 2004; Yoda 2005, Dickins 2009, also cf. den Dikken 2005)[1], the type of agreement in (2) is a long agreement, and we can account for it by adopting an Agree across phases (AAP) mechanism, a type of long-distance agreement (LDA) (see e.g. Bhatt and Keine 2017).

In this study, we tackle this topic for several reasons: i) YIA seems to provide interesting data for wh-agreement, which is found only in a small number of languages such as Greek, Indonesian, Lubukusu. A wh-agreement is a phenomenon in which the wh-word agrees with the complementizer, verb (and subject). It is as an agreement morphologically marked on either the verbs or complementizers or both (see e.g. Lochbihler and Mathieu 2016; Jeoung 2020), ii) to the best of our knowledge, no study has tackled this phenomenon in YIA (nor in any Arabic variety, if any), iii) the data presented and discussed in this study would open several windows that would attract Arab and NonArab linguists to investigate YIA or Yemeni dialects further, and iv) YIA is very under-investigated language variety like other Yemeni Arabic vernaculars, having received little linguistic attention. Although a few Yemeni dialects, such as San'ani, Adeni and Hadrami, have been to some extent researched, YIA dialect has been almost entirely neglected until recently (see e.g. Shormani 2019; Shormani and Al Hussen 2024). Thus, in this study we investigate wh-agreement in YIA adopting a long-distance agreement approach anchored Feature Inheritance (FI), arguing in favor of Chomsky (2007, 2008) that T/V inherits its agreement feature from C, given the agreement manifested between Wh-Op(erator)/Sub(ject), C and V (cf. also Carstens 2005; Boeckx 2004; Bhatt and Keine 2017).

The rest of this article is arranged as follows. In section 2, we outline the theoretical foundations. In section 3, we spell out our proposal. In section 4, we outline the grammaticalization of third person pronouns in YIA. In section 5, we examine wh-agreement in simplex and complex wh-constructions including cyclic/multiple wh-movement. In section 6, we apply our proposal, specifically AAP to wh-agreement across languages. Section 7 concludes the article providing some implications for minimalist syntax, UG and NLP.

[1] See Footnote (12) for more details on *li-*

## 2. Theoretical foundations

In this study, we adopt several core minimalist notions, including Phase Theory, Core Syntax within which several notions such as Agree, Long-distance Agreement, Feature Inheritance, among other relevant concepts will occur. These are outlined in the following (sub)sections.

### 2.1. Phase Theory

Phase Theory (PT) represents one of the most influential developments in current minimalist syntax. Its central assumption is that syntactic derivation proceeds cyclically in phases rather than as a single chunk/computation (see e.g. Chomsky 2001, 2008). This architecture is motivated by the view that the human computational system, precisely $C_{HL}$, cannot process large syntactic structures all at once, but instead constructs, computes and interprets them incrementally (Chomsky 2001, 2008). A phase can thus be understood as a syntactically complete and semantically interpretable chunk that is transferred to the interfaces during the course of the derivation.

A phase is generally considered semantically complete when it constitutes a coherent unit of interpretation that can be interpreted independently of larger structures. This conception is closely related to the idea of cyclic transfer and, in some approaches, Multiple Spell-outs. Building on Chomsky's work, Gallego (2012: 5) characterizes phases as syntactic objects that are (see also Bošković 2014, 2016):

(3) a. as small as possible

b. capable of receiving at least some reasonable interpretation at the C-I interface

c. identifiable in a natural and straightforward manner.

Another problem PT provides solution to is manner and place of Spell-out. Along these lines, Panagiotidis (2010: 2) argues that PT offers a solution to "an important problem of theoretical nature: how best to conceive the manner and place of Spell-out in the syntactic derivation."

Within the phase-based architecture, Chomsky (2001, *et seq*) argues that a phase is a unit of syntactic computation that can be transferred to the interfaces as an independent syntactic object. In this sense, phases constitute what Chomsky (2001: 1) describes as "an optimal solution to minimal design specifications." Traditionally, two syntactic domains have been identified as phases: CP and transitive vP (v*P), though subsequent research advocate for other categories, such as DP and PP, to exhibit phase-like properties (see e.g. Shormani 2016). One defining desideratum of phases is that, once transferred, their internal structure becomes inaccessible to further syntactic operations. Only the edge of the phase remains available for subsequent computation. This restriction is formalized as the Phase Impenetrability Condition (PIC) (Chomsky 2000: 108). It follows that a phase is complete when all relevant feature-checking/valuation operations have been satisfied, including the valuation of unvalued features and the elimination of uninterpretable features. Once these requirements are met, the phase, viz., CP/vP is transferred to the interfaces, namely LF and PF. The introduction of phases addresses a fundamental question concerning the architecture of grammar, namely the timing and location of Spell-out in syntactic derivation (see also Panagiotidis 2010). It follows that phases correspond to those syntactic domains in which essential feature valuation processes take place, particularly the valuation of structural Case and unvalued φ-features. It also follows that tensed clauses and transitive verb phrases (v*P) qualify as canonical phase domains. Chomsky (2008) further argues that the relevant clausal phase is CP rather than TP.

As for wh-movement, PT also has important implications for it and for all types of movement operations. For example, an XP (precisely Wh-Op word/phrase) undergoing A'-movement cannot remain at the V-compl configuration, specifically in non-in-situ wh-languages. Instead, it has to move successively through CP edges until it reaches its final landing site, typically the left periphery of CP (see e.g. Citko 2014; Keine and Zeijlstra 2025). This phenomenon, known as successive-cyclic movement, follows naturally from the PIC, since movement has to proceed through accessible phase edges rather than directly from the inaccessible interior of a phase (cf. Boeckx and Grohmann 2007). Although the interior, i.e. head and its complement, of a transferred phase is not accessible for further syntactic operations, it can still participate indirectly in later computations. Chomsky (2008) discusses cases in which material inside a lower phase contributes to the valuation of features on a higher *probe*. One well-known example involves the inherent nominative object of an experiencer verb, which may determine the φ-features of a higher T, despite being contained within a lower phase (see Sigurðsson and Holmberg 2008; Chomsky 2008). Such cases illustrate that phase boundaries constrain syntactic operations without completely isolating transferred material from the computational system.

To elaborate, after its introduction stage, PT has been developed and some notions were even refined by linguists including Chomsky himself. One of the most important notions that have been refined is PIC, which states that only the phase edge can be accessed by further computations. However, linguists including (Bošković 2007; Boeckx 2009a & b; Roberts 2010) provide evidence from several languages such as Hindi, Icelandic, Tsez, Basque, etc. in which Agree can sometimes affect phase heads. Put differently, the interior of a phase may still be accessible for certain interpretive purposes, such as determining the valuation of a higher probe (Chomsky 2008). This empirical evidence makes linguists even Chomsky himself refines the notion of PIC (see also Rouveret 2008). Chomsky (2008) proposes that Agree taking place between phases is not subject to PIC effects, as these structures give rise to no problem. Chomsky (2008) argues that, given strong empirical evidence, PIC "holds only for the mappings to the interface, with the effects for narrow syntax automatic" (p.143). The empirical evidence was brought from Icelandic where *Agree* takes place "into a lower phase without intervention in experiencer constructions in which the subject is raised (voiding the intervention effect) and agreement holds with the nominative object of the lower phase" (p. 159). Shormani (2017) also provides strong support to this phenomenon from YA, where Agree "looks down" into the search space for a very far away goal (see also Rouveret 2016).

## 2.2. The core syntax

The notion *core/narrow syntax* refers to how $C_{HL}$ constructs/computes a piece of language via the mental operations, namely *Select, Merge, Agree* and *Spell-out.* In minimalism, specifically PT, this depends heavily on LIs' features, as they play an important role in the derivation by triggering these mental operations. For example, the operation *Select* selects a LI from the lexicon/numeration and "hands" it over to Merge, which in turn merges/combines an LI feature K with another LI feature L. This merging results in an ordered pair X(K,L), in which X$^{o}$ results lexical head. And depending on K/L's features, Agree comes to place. Features include φ- (number, gender, person), WH, Case, (in)def, which are on both functional heads like C, T, v, and lexical DPs, *probes* and *goals*, respectively (Chomsky 2000, *et seq*; Boeckx 2003). As noted above, features on *probes* are usually unvalued/uninterpretable, and those on goals are usually valued/interpretable. For the derivation not to crash, the unvalued/uninterpretable features must be valued/uninterpreted before the whole chunk is sent to the interfaces.

### 2.2.1. Agree

Feature valuation/interpretation is carried out via Agree. Agree results in some sort of MATCHING, i.e. a suboperation called MATCHING results from Agree. When Agree takes place, all unvalued/uninterpretable get valued/uninterpreted and then deleted. However, some features cannot be satisfied by Agree such as WH, EPP, which require an IM operation from their in-situ positions to Spec,CP and TP, respectively. When all features are valued, the chunk gets ready to transfer to the interfaces, and then spelled out by Spell-out operation. This chunk is called a phase, and all this interaction takes place within this phase, as just noted.

Thus, since features trigger (process and compute) these mental operations, and since we are dealing with wh-question phenomenon and the operations it contains, Feature Triggering could be formulated in (4) (cf. Chomsky 2000; Boeckx 2003; Shormani 2017: 151).

(4) a. Features trigger MATCHING (e.g. there is a valued/interpretable [WH] feature on the wh-word that matches the unvalued/uninterpretable [WH] feature on Cs).

b. Features trigger IM (e.g. i) Wh-Op remerges in Spec,CP, ii), V in C (through V & T), and iii) Subj-DP in Spec,TP.

c. Features trigger Agree (e.g. the value(s) of goal features (e.g. Wh-Op, DP, *pro*) match those of the probe ($C^o$/$T^o$).

### 2.2.2. Feature inheritance

Questions such as: where do features (e.g. those in 4 above.) come from?, Are they part of functional heads e.g. T, v?, Do they inherit these heads within the TP-domain?, and/or Are they inherited to some higher functional head in CP-domain, or otherwise the information-structure, and then transmitted onto TP heads? These questions, among related to others, were the most overwhelming in post-phase era. Scholars such as Rizzi (2004) concluded that these features are not in-built in T or v, but rather they are inherited to some head(s) in the left periphery, specifically CP-domain. They based their conclusions on evidence data from across languages such Icelandic, (see e.g. Sigurðsson and Maling 2010), Greek (see e.g. Panagiotidis 2010), Arabic (see e.g. Shormani 2017), among other scholars and languages (see also Richards 2012). A central issue in post-phase minimalist syntax concerns the locus and distribution of formal features. Central questions include those we posed above, becoming particularly pressing in the post-phase era leading to the development of the feature inheritance model.

Within Chomsky's (2005) and later work, features are not freely distributed across all functional heads. He proposes that only phase heads, namely $C^0$ and $v^{*0}$, are the true loci of formal features. Non-phase heads such as T and V do not inherently carry full feature specifications; rather, they receive them derivatively through a mechanism of FI. In this view, C acts as a phase head that may host agreement-related features and subsequently transfers them to T in the narrow syntax once T is merged. The mechanism regulating in this distribution is what Chomsky (2005, 2008) terms as Feature Inheritance, which specifies how features are transmitted from phase heads, viz., $C^0$, $v^{*0}$ to non-phase heads, T, V. This operation can be schematized in (5).

(5) [$_{CP}$ C{φ} [$_{TP}$ T [$_{vP}$ … ]]] → [$_{CP}$ C [$_{TP}$ T{φ} [$_{vP}$ … ]]]

In the narrow syntax, T inherits Agree features from C, as phase heads regulate the distribution of unvalued features across the derivation. Building on this framework, the present analysis adopts FI approach and assumes that T receives its φ-features from C in the course of derivation. This is consistent with Chomsky's (2005, 2008) claim that phase heads are the primary loci of feature specification. Given antecedent relations in the derivation, it can be argued that the "antecedent" referred to in Chomsky (2008) corresponds to an A'-domain element, which in our analysis aligns with the wh-Op (cf. Shibata and Yashima 2014' Rouveret 2015). This is supported by data where agreement relations extend across the A'-domain, involving Wh-Op, relative operator, and verb agreement.

**(6) Feature Inheritance**

uF must spread from edge to non-edge (i.e., from C to T, v* to V, etc.)

(Richards 2008: 567)

Additionally, certain languages provide further empirical support for treating tense as a CP-dependent feature. For instance, in Irish C has been argued to encode tense distinctions, including past versus non-past morphology (Adger 2007: 34). Such evidence reinforces the view that tense is not inherently associated with T, but is instead introduced and valued in the CP domain and subsequently inherited by T. This antecedent factor is at the heart of the proposal pursued here: assuming for the moment that the notion 'antecedent' implied in Chomsky (2008) refers to a constituent in the A`-dependency domain (cf. Rouveret 2008, 2015), which could be the Wh-Op in our system.

However, in YIA there are data that cast doubt on FI in (6). Consider (7) a 2 CPs wh-question, in which C1 agrees with some C2, T2, v2, and *pro*, and not with all.

(7) a. man-u qāl ʕali ʔinn-uh jāʔ ?
who-3MS said.3MS Ali that-3MS came-3MS
'*Who did Ali say that he came?'

b. man-eh qāl ʕali ʔinn-eh jāʔ-ah ?
who-3FS said.3MS Ali that-3FS came-3FS
'*Who did Ali say that she came?

c. man-en qāl-ah ʕalia ʔinn-en jāʔ-ayn ?
who-3FPL said-3FS Aliya that-3FPL came-3FPL
'*Who did Alia say that they came?'

[$_{CP1}$ man-en [C ] [$_{TP1}$ [T qāl-ah] [ $_{vP1}$ ʕalia [ v ~~qāl-ah~~] [ $_{VP1}$ [V ~~qāl-ah]~~ [$_{CP2}$ [~~man-en]~~ [C ʔinn-en] [TP2 [T jāʔ-ayn] [ $_{vP2}$ ~~[man-en]~~ [ v ~~jāʔ-ayn~~] [$_{VP2}$ [V jāʔ-ayn]]]]]]]]]

Example (7a) supports (6) because C1 seems to agree with T1, v1, C2, T2, v2; all are 3$^{rd}$ person, masculine and singular. However, in (7b & c) the wh-words or C1 *man-eh, man-en* agree with C2, T2 and V2 *ʔinn-eh, ʔinn-en* and *jāʔa-ah, jāʔ-ayn*, *pro,* respectively. However, T1/V1 does not agree with C1; rather it agrees with the DP-subj *ʕali/ʕalia* in both examples.

Similar problems were touched on by Chomsky (2005: 18), leading him to explicitly suggests that C "may be the locus of agreement, selecting T and assigning it (unvalued) φ-features." This position is further reinforced by him arguing that "T, φ-features and Tense appear to be derivative, not inherent," since tense and tense-like properties (e.g., irrealis) are fundamentally determined at the CP level" (Chomsky (2008: 143).[2] This suggests that φ-features and tense

[2] Realis refers to mood used for events or states that are understood to be actualized, factual, or currently occurring. However, irrealis mood encompasses everything that is not yet a reality such as hypothetical scenarios, future

are not intrinsic properties of T; they may depend on C, based on the syntactic environment they occur in. The fact that C1 in (7b & c) agrees with C2, T2, V2, but not with T1/V1 strongly supports Chomksy's argument. Given these empirical facts, we propose (6') as a refinement of (6).

**(6') Feature Inheritance**

Let $\alpha_i \in \{C, v, (n)\}$ be phase head, and β a nonphase head such that $uF(\beta) \subseteq uF(\alpha_i)$

(cf. Chomsky 2008: 143)

In (6'), α represents phase heads, specifically C, v, (*n*), identified as the primary loci where uninterpretable features *uF* originate. β, however, denotes a non-phase head, precisely T/V whose *uFs* are a subset of C's ones. *N* refers to any other phase heads such as K of the Construct State (CS) see Shormani 2016), APs (Al-Rawi 2019). It follows that a nonphase head may select all or a subset of C's features depending on the syntactic environment it occurs in. Thus, in (6') β acquires only the features from C that meet its syntactic environment. The syntactic environments of (7b & c) "compels" T to agree with its subject, namely *ʕali* and *ʕalia,* respectively. Our refined form of (6) seems to be scientifically sound; it has a wider scope and accounts for the data straightforwardly (cf. Aoun et al. 1981). Thus, in some environments T inherits a subset of its *uF* from C as in (7a & b) and some other times all the features as in (7a).[3] This subset may include Case, some φ-features (e.g. person), which unveil that T partly (but not completely) inherits its features from C.

### 2.3. Wh-movement: How far can a Wh-Op move?

In this section, we sketch wh-movement in the literature, attempting to answer one major question: How far can a Wh-Op move? Our answer considers cross linguistic views and

---

projections, commands, desires, or counterfactuals. The difference between the 2 terms is characterized as a feature-based variation in the functional architecture of the clause, specifically within the MoodP or T head. A realis structure is typically "marked" by the projection of a full, finite CP, allowing for standard T-to-C movement and the licensing of temporal features. Conversely, an irrealis structure often reflects a truncated syntactic tree or an underspecified T-node, where the absence of finite features blocks standard temporal anchoring and instead triggers a null modal operator in the left periphery, effectively shifting the clause from a factual, time-bound representation to a hypothetical, non-finite state (see e.g. Chomsky 2008; Rouveret 2008).

[3] See also Shormani (2017: 147/8) who provides empirical evidence for FI from SA including (i).

(i) a. ṭ-ṭaalibu llaði jaaʔ-a pro ʔams-i.
the student.3MS who.3MS came-3MS yesterday-GEN
'The student who came yesterday.'

b. ṭ-ṭullaabu llað-iina jaaʔ-uu pro ʔams-i.
the-student.3MPL who-3MPL came-3MPL yesterday-GEN
'The students who came yesterday.'

c. qaabal-tu ṭ-ṭaalibata llati jaaʔ-at pro ʔams-i.
met-I the student.3FS who.3FS came-3FS yesterday-GEN
'I met the (f) student who came yesterday.'

In (ia-d), C *llaði, llað-iina, llati* and *llāti,* respectively) agrees with the constituent it introduces, namely *r-rajul*, *r-rijālu*, *al-bintu* and *al-banātu*, respectively, in all φ-features.

Shormani and Al Hussen (2024) also provide strong evidence that C has also Case. This is evidenced in (ii).

(ii) a. ʔallah-u yaʕlamul-ħaqiqat-a
God-NOM knows the-truth-ACC
'God knows the truth.'

b. ʔinna ʔallah-a yaʕlamul-ħaqiqat-a
C God-ACC knows the-truth-ACC
'Indeed, God knows the situation.'

In (iia), the topic *ʔallah* carries a (default) nominative Case, but in (iib), it carries an accusative Case, which explicitly signals that the C *ʔinna* assigns Case, precisely accusative.

provide robust empirical evidence from YIA that the Wh-Op has to stop in Spec,VP before proceeding to higher layers. Wh-movement is a syntactic process where Wh-Op/word is moved to the left periphery, precisely Spec,CP in the core syntax. Ross (1967) is considered the pioneering linguist who examined the restrictions governing extraction processes. He demonstrated that wh-elements cannot move freely from any position within a sentence and identified several structural domains that block extraction. These constraints, known collectively as island constraints, include the *Complex Noun Phrase Constraint, Coordinate Structure Constraint,* and *Subject Condition* (see also Aoun and Li 2003). Ross's work established the notion that movement operations are subject to locality restrictions and laid the foundations for subsequent research on syntactic dependencies. For instance, Chomsky (1977) argues that wh-movement is a universal syntactic process governed by UG principles.

Building on his 1977 work, Chomsky (1986) developed a more formal account of wh-movement within the P&P framework. In this model, a wh-phrase moves from their base-generated/merging positions to the Spec of CP in order to satisfy the requirements of interrogative structures. Chomsky proposed that movement leaves a trace behind, creating a chain that links the moved constituent to its original position. The theory of Subjacency was introduced to explain why certain extractions are grammatical while others violate locality conditions (Aoun and Li 2003). In this view, wh-movement is constrained by structural principles that limit the distance and manner in which constituents may move, roughly schematized in (8).

(8) $\textbf{Wh}_{\textbf{-movement}}$ = [$_{CP}$ [Wh$_i$ [Base _t$_i$_]]] ➔ *L*

where CP [Wh$_i$ is the landing site, [Base is the base-generated position, $t_i$ (trace), the start of wh-chain, ➔ iff *L* {Locality Conditions (LCs)}

(8) witnessed several refinements in the analysis of wh-movement phenomena, where the main concern was what exactly constitutes LCs. For example, Koopman and Sportiche (1991) argue that a Wh-Op (NP object) has to move to Spec,VP for Case requirements, providing evidence from French (and some other languages in support of their arguments). In their own words, "the verb assigns structural Case to its Spec, we get an immediate account for the position in which they occur… An NP that needs structural Case moves from its thematic position to [Spec, VP] because it has to satisfy the Case filter; only one NP may move there, because X-bar theory only makes one Spec position available." (p. 64).

Larson (1988), addressing double object constructions, provides structural innovations that influenced subsequent treatments of movement and extraction inclining wh-movement, specifically "superiority" and c-command notions where "[a] wh-phrase cannot in general be moved over another wh-phrase that c-commands it (in other words, is "superior" to it) in underlying representation" (p.336). [4]

---

[4] Following Koopman and Sportiche (1991), A'-positions need not be restricted to Spec,CP. They concluded that "[a]djoined positions and [Spec, CP] are A-bar positions

1. [Spec, VP] and/or [Spec, $V^{max}$] are A-positions
2. [Spec, IP) can be an A-position (English, French, Arabic) and, possibly, an A-bar position (English, Yiddish, Malayo-Polynesian)
3. [Spec, ASPP] can be an A-position (English, French) or an A-bar position (Kilega)" (p.77).

They independently argue that preposed objects in French pass through Spec,VP during wh-movement. Thus, these accounts provide a theoretical precedent for treating projections within the verbal domain as intermediate stages in successive-cyclic movement. We extend this line of reasoning by proposing that, in IYA, Spec,VP itself functions as the relevant intermediate A'-landing site for extracted wh-objects before movement to the CP domain (see also Sportiche 1988).

However, Minimalism raises a substantial question on whether the wh-movement undergoes a "one-swap" movement, specifically in multiple wh-movement constructions from its base position to its (final) landing site, or through in-between "other Specs". Given that wh-movement is an A'-movement, it can only move to non-argument position. However, some scholars propose multiple specifiers where a wh-word can move to in its journey. Given PIC, Radford (2009) proposes that since vP is a phase, it could necessitate a successive-cyclic approach to wh-movement to comply with it (see also Woods 2012). Woods develops Rudin's (1988) seminal pre-minimalist argument that in multiple wh-questions, there is no superiority effect, proposing that Spec,vP (or even multiple Specs in some other works) may be a possible site for wh-word movement before reaching its landing site. However, this proposal may not be on the right track as it violates some UG principles including A-argument position, theta theoretic notions, and subjecthood (see also Larson 1988).

In YIA, there seems to be a different matter, as wh-extraction in this language poses a significant challenge to standard PT. While many current approaches assume that wh-arguments must stop at the edge of the vP phase, the data from YIA suggest a different story. Specifically, the presence of overt object clitics (e.g. *-hen* in *nāwal-hen*) together with the floating quantifier *kull-hen* 'all of them' in long-distance wh-extractions indicates that the extracted object maintains a close morphological and syntactic relation with the verb even after movement towards CP has begun (cf. Boeckx 2003; Gallego 2017). The floating quantifier further shows that part of the nominal chain remains interpreted within the VP domain, providing independent evidence that movement proceeds through an intermediate VP-internal position rather than directly through phase edges. This raises several important questions: Why does the object interact so closely with the verb at a low structural level? How does this interaction avoid the intervention effects normally created by the subject in Spec,vP? And why is the floating quantifier licensed precisely in the VP domain?

To answer these questions and for (9) to satisfy locality conditions, we propose that argumental wh-phrases first move to Spec,VP. Consider (9), with its schematic representation underneath.

(9) ʔayyan kutub qāl ʕali ʔinn-uh nāwal-hen kull-hen li-lʕyāl?
which books said.3MS Ali that-he gave-3MS all of them to-the.boys
'Which books did Ali say he gave them all to the boys?'

[CP1 ʔayyan kutub [C [TP [T qāl [ vP1 ʕali [v [~~qāl~~] [VP1 [V [~~qāl~~] [CP2 [~~ʔayyan kutub~~] [C ʔinn-uh [TP2 ~~ʕali~~ [T nāwal-hen [vP2 ~~ʕali~~ [v ~~nāwal-hen~~ [VP kull-hen [V ~~nāwal-hen~~ [DP [~~kull-hen~~] [DP [~~ʔayyan kutub~~] [PP [li-lʕyāl]]]]]

Keeping (4), (6′), and (8) in mind, the derivation in (9) proceeds through successive-cyclic movement, where the argumental wh-phrase *ʔayyan kutub* has to satisfy locality conditions. We show that the wh-object does not simply target CP phase edges; rather, it undergoes a mandatory VP-internal licensing operation motivated by the following UG principles: i) VP-internal licensing. The wh-argument *ʔayyan kutub* is externally merged as the complement of the verb *nāwal* 'gave'. Before successive-cyclic movement proceeds, it undergoes internal movement to Spec,VP, where it establishes a local licensing relation with the lexical verb within the VP domain. Following Woolford's (2006) view that lexical licensing of internal arguments is confined to the VP proper, we propose that this VP-internal configuration constitutes the necessary licensing stage preceding extraction.[5] The object clitic *-hen*

[5] Woolford (2006) proposes that lexical Case is licensed by lexical heads such as V within the VP domain, whereas inherent Case is licensed by little v and is associated with external arguments and certain goal arguments. This architecture reinforces the view that the VP constitutes the primary syntactic domain in which internal arguments establish local licensing relations with the lexical verb before participating in higher syntactic operations. In the

morphologically realizes this local dependency, while the stranded floating quantifier *kull-hen* independently diagnoses the persistence of the lower copy within the VP domain (see also Ott 2015). We argue that this VP-internal licensing relation has to be established before the derivation proceeds to higher phase edges, ii) Subject-bypass and locality. By first moving to Spec,VP, the wh-phrase establishes its VP-internal licensing relation before the external argument *ʕali* is merged in Spec,vP. It follows that once the subject enters the derivation, the wh-object has already exited the complement domain of the verb, thereby preventing the subject from intervening in the dependency between the verb and the extracted object. This derivational ordering eliminates potential locality violations and allows subsequent successive-cyclic movement through the higher CP phases without infringing Relativized Minimality (Rizzi 1990), the Shortest Move Constraint, or the Minimal Link Condition (Chomsky 1995; Müller 2011; Fischer 2018)[6], iii) Feature valuation and convergence. From Spec,VP, the wh-phrase moves successively through Spec,$CP_2$ and finally Spec,$CP_1$. At each stage, the [uWH] feature on C is valued by remerging wh-phrase. The lower copy is morphologically realized by the resumptive object clitic *-hen* on *nāwal-hen*, while the floating quantifier *kull-hen* provides further overt evidence for the intermediate position occupied by the moving DP.[7]

Within the Copy Theory of Movement (Chomsky 1995; see also Chomsky 2005, 2007, 2013), these overt elements preserve the morphological reflexes of the movement chain, allowing the derivation to converge at PF while maintaining the interpretation of the displaced argument, and iv) Argument-adjunct asymmetry. We propose that this type of wh-movement is restricted to argumental wh-phrases. Unlike adjunct wh-elements, argumental DPs enter into a local morphosyntactic dependency with the lexical verb, a dependency independently diagnosed by object cliticization and floating-quantifier stranding. We therefore propose that only argumental wh-phrases are required to pass through Spec,VP before successive-cyclic A'-movement proceeds (cf. also Doliana 2022). Adjuncts such as *mata* 'when' and *kayfa* 'how', which neither participate in verb-argument licensing nor exhibit clitic doubling or floating-quantifier licensing, are exempt from this requirement and may therefore move directly to the CP domain, precisely Spec,CP (but see Koopman and Sportiche 1991; Rizzi 1990). The argument-adjunct asymmetry thus follows naturally from the distinct licensing requirements imposed on arguments and adjuncts, rather than from differences in the mechanics of wh-movement itself (cf. also Reeve and Hicks 2017; Keine and Zeijlstra 2025).

In our analysis, Spec,VP is not simply an "escape hatch" but a mandatory VP-internal licensing position required for argumental wh-phrases. After moving to Spec,VP, the extracted object establishes a local morphosyntactic dependency with the lexical verb before successive-cyclic movement proceeds. This VP-internal configuration is associated with two distinct morphological reflexes: i) the object clitic *-hen* on *nāwal-hen* is taken as a result of Agree as MATCHING outlined in (4) (see also Revourt 2008) realized the "tie" between the lexical verb and the internal argument, thereby encoding the local verb–object dependency established in Spec,VP prior to movement into higher domains (see e.g. Abels and Dayal 2023), and ii) the present analysis, we extend this insight by proposing that in YIA, argumental wh-phrases obligatorily pass through Spec,VP to establish this local morphosyntactic dependency before successive-cyclic movement proceeds.

[6] Following Lakoff's (1971) notion of a "global rule", Müller (2011: 13) advocates that "the relative order of quantified items that is present at the beginning of the derivation must be maintained throughout later levels of the derivation, even under quantifier movement". Müller's argument stems from the fact that "order preservation effects observable with quantifier lowering (where the relative scope of quantified items corresponds to their surface order) can be accounted for by a "global rule" (see also Müller 2001; Williams 2003; Bobaljik and Wurmbrand 2008).

[7] Put differently, the resumptive object clitic *-hen* may be analyzed as the morphological realization of the lower segment of the movement dependency, while the stranded floating quantifier *kull-hen* provides independent evidence that part of the dependency remains within the VP domain.

floating quantifier *kull-hen* provides evidence for an internal copy of the moved wh-phrase stranded within the VP domain, in line with the Copy Theory of Movement (Chomsky 1995, 2005, 2007).

To recapitulate, since the object reaches Spec,VP before the external argument is merged in Spec,vP, it has already exited the complement domain when the subject enters the derivation. It then follows that the subject does not intervene in the movement dependency, thereby avoiding potential violations of Relativized Minimality, the Minimal Link Condition, and related locality constraints. This analysis also captures the asymmetry between argumental and adjunct wh-elements. Only argumental wh-phrases are required to pass through Spec,VP because they must first establish a local licensing relation with the lexical verb before successive-cyclic A'-movement proceeds. By contrast, adjunct wh-elements such as *mata* 'when' and *kayfa* 'how', which are not selected by the lexical verb and neither participate in VP-internal Agree relations nor exhibit clitic doubling or floating-quantifier diagnostics, are exempt from this requirement and may therefore move directly to the CP domain. We therefore argue that Spec,VP constitutes a language-specific licensing position rather than a conventional phase edge. By relocating the crucial intermediate step from Spec,vP to Spec,VP, the proposed analysis simultaneously accounts for the distribution of resumptive object clitics, the behavior of stranded floating quantifiers, and the interaction between VP-internal licensing and locality in YIA (cf. also Shlonsky 1991; Krivochen 2021). It therefore provides a unified account of why argumental wh-phrases obligatorily undergo VP-internal licensing prior to successive-cyclic movement, whereas adjunct wh-elements proceed directly to the CP domain.[8]

---

[8] Note that our proposal departs from Luigi Rizzi's (1990) analysis in both its empirical motivation and theoretical implementation. While Rizzi treats Spec,VP primarily as an intermediate A'-position in French movement constructions, we argue that in YIA Spec,VP constitutes a mandatory VP-internal licensing position through which argumental wh-phrases must pass before successive-cyclic movement proceeds. However, several observations in Rizzi's account provide independent support for the availability of such a position within UG: i) in his analysis of French *combien* 'how many/how much' constructions, Rizzi proposes that Spec,VP serves as an intermediate landing site in the derivation. Although this position is not θ-governed, it is antecedent-governed and therefore satisfies the licensing conditions on movement without violating Relativized Minimality, despite the presence of intervening elements such as *pas* (Rizzi 1990: 18). This observation establishes that Spec,VP may legitimately function as an A'-position involved in movement dependencies, ii) discussing floating quantifiers, Rizzi argues that the hypothesis that *beaucoup* occupies the A′-specifier of VP is compatible with Subject-in-VP approaches (see e.g. Sportiche 1988), according to which the thematic subject originates within the VP domain. He notes that the rigid ordering in (i) supports a structural hierarchy in which the VP-adjoined subject occupies a higher position than the A′-specifier of VP.

(i) Ils ont tous beaucoup mangé.
'They have all eaten a lot.'

Rizzi further observes that the leftward-moved *tout/tous* proposed by Kayne (1985) likewise occupies Spec,VP. The incompatibility of *tout* with *beaucoup* and the fact that it follows the subject-floated quantifier *tous* provide additional evidence for the existence of this distinct VP-internal position, consider also (ii).

(ii) Ils ont tous tout mangé.
'They have all eaten everything.'

And iii) and most relevant to the present analysis, Rizzi reviews the proposal of Sportiche (1988), who argue that direct objects move to Spec,VP in order to receive structural Case before undergoing further movement, perhaps along the lines proposed by Larson (1988). Subsequent adjunction to VP is independently motivated by definiteness-related considerations, while the trace left by this additional movement accounts for the degraded acceptability of certain extraction structures (Rizzi 1990:128). Although our analysis differs in treating Spec,VP as an obligatory stage of long-distance argument extraction in YIA, rather than a language-specific response to French quantifier placement or definiteness effects, these proposals collectively demonstrate that Spec,VP has independently been recognized as a legitimate locus for A'-movement, Case-related operations, and locality-sensitive dependencies. We therefore view our proposal not as introducing a novel structural position, but as

## 3. Proposal

Given $C_{HL}$ and its ability of generation, a wh-agreement construction can be as long as possible. Consider (10), a YIA multiple wh-question undergoing several wh-movements.

(10) ʔayyan kitāb qāla ʕali ʔinn-uh ʔaddy-uh liman dār ħuwār byyn-uh wa
Which book said.3MS Ali that-he gave-he to whom had a talk between-him and
bayn muhammad ʕann-uh?
between Mohammed about it
'Which book did Ali say that he gave to who had a ward between him and Mohammed about it"

Such multiple wh-questions in which each wh-word/phrase undergoes multiple wh-movements from their in-situ positions to their final landing sites require a mechanism in which Agree operates. The traditional form of Agree may not account for such structures due to containing multiple phases having several probes, Cs, Ts, vs, along with IPC. With this in mind, we propose that wh-agreement in YIA takes place within Agree across phases, which is a type of long-distance agreement.

AAP is a phenomenon where a probe such as C/T probes downwards for a goal with a valued feature in its search space "till finding" this valued/interpretable feature in "a very far away place". Then a AAP is established between both, the result of which is valuing and interpreting all unvalued/uninterpretable features. However, when C/T, for example, probes downwards in YIA wh-questions, it will find several probes in-between till reaching the target goal. As a solution out of this problem, we propose that the topmost C1 in (12) constitutes a MATCHING Agree with all the probes downwards, and when AAP is established between it and some goal, all the unvalued/uninterpretable features are valued/interpretable (cf. Frampton and Gutmann 2006; Pesetsky and Torrego 2007). This is schematized in (12).

**Agree Across Phases**

(11) Given (6'), in a multiple wh-movement question,

i. Agree as MATCHING (AM): α probes for a γ with corresponding valued/interpretable features in its Γ chain. This relation establishes a copy/MATCHING-link between α and all instances of the chain $\Gamma = \{\gamma 1, \gamma 2, \ldots \gamma_n\}$. $AM(\alpha, \Gamma) \Leftrightarrow F(\alpha) \cap F(\Gamma) \neq \emptyset$.
ii. Feature valuation (FV): FV requires a local licensing/valuation relation. FV(β, γ) occurs only if γ is local to β and $\exists uF(\gamma)$. $FV(\beta, \gamma) \rightarrow V(\gamma) = vF \Longrightarrow Active(\gamma) = 0$
iii. Nonphase transparency (NT): $\forall \beta \notin \alpha$, $Intervener(\beta) = \emptyset$. βs facilitate AM but do not block FV for higher probes.

The basic idea in (11) is that T/V's FV is not linked to AM. This notion is basically based on "Separate Probes" (see e.g. Heck and Richards 2010; Branan and Erlewine 2024). Given this, there are 3 axes in (11): i) AM serves as the initial mechanism for establishing a long-distance relationship. Defining MATCHING as the non-empty intersection of features $F(\beta) \cap F(\gamma) \neq \emptyset$, the probe is permitted to establish a formal copy-link with an entire chain, precisely Γ. This is necessary for multiple wh-movement constructions, where a single probe has to account for the distributed features of multiple elements (cf. Hein and Murphy 2020). To decouple the matching operation from the strict, local phase boundary, our system allows the probe to extending an independently motivated one to account for the interaction between object cliticization, floating-quantifier stranding, and successive-cyclic wh-movement in YIA (cf. also Keine and Zeijlstra 2025). The morphological evidence provided by the resumptive object clitic and the stranded floating quantifier further distinguishes our analysis by supplying overt diagnostics for the VP-internal licensing relation, evidence that is absent from the French constructions discussed by Rizzi.

identify relevant targets across the derivation regardless of their hierarchical distance, ii) this workflow maintains the integrity of the minimalist link for local licensing through the FV condition. Although MATCHING may occur nonlocally, the subsequent FV is strictly constrained to a local configuration. This bifurcation is logically necessary to prevent overgeneration; it ensures that while a probe may identify a distant target, the final assignment of values remains subject to the locality conditions essential for interface interpretability (see also Miyagawa 2009; Hein 2024; Abels and Dayal 2023).

And iii) our system ensures derivational efficiency through the deactivation of goals and the NT condition. The stipulation that FV triggers the deactivation of the goal (Active(γ) = 0) prevents redundant operations and potential valuation conflicts, thereby maintaining economy. Thus, the NT condition establishes a formal “pathway” for the probe. By permitting nonphase heads to act as transparent non-interveners (cf. Miyagawa 2009; Hein 2024), the derivation effectively bypasses rigid phase boundaries even without violating the PIC, thereby facilitating long-distance dependencies while adhering to the economy of C<sub>HL</sub> (see also Richards 2012; Cheng 2025). Thus, we can conclude that AM is both discourse and syntactic in nature, while FV is syntactic *per se.* Discoursely, the former takes place across phases, which is the essence of AAP; it interprets contents of the (A’)-chains like antecedents, copies, *pros* (and null categories) (cf. e.g. Ouhalla 1999). Syntactically, it works with FV in several phenomena including RP articulation, Case spell out, etc. FV, on the other hand, focuses mainly on syntax-internal phenomena like valuing φ-features.

Simply put, in (11 (given 6’)), once C/v is valued for a feature F, it will enter an AM relation with all the lower instances of C/v, and T/V will not intervene in this MATCHING relation. In our context, C1/v1, C2/v2, C3/v3, Cn/vn; let F have 2 values: *u*F and *v*F. Let also F be [WHs]; Case; φ, and FV requires a local licensing/valuation relation, FV(T, DP) occurs exclusively if DP, having vF is local to T, which has uFs. Once valuation is complete, the DP is deactivated; it will not intervene/block next AM/FV processes such as AM between Cs/vs and Wh-Op in the MATCHING chain. It follows that once the Wh-Op is (re)merged, all the instances of *u*F will be valued via AAP schematized in (12).

(12)[CP1 C1[uWH] [uφ] [vCase [TP T1[uφ][vCase]…[vP1 Subj/pro [v][vφ] [uCase] [VP1 [V1inf]…[CP2 C2[UWH]….Wh-Op [[vWH;vφ; uCase]]]]]]

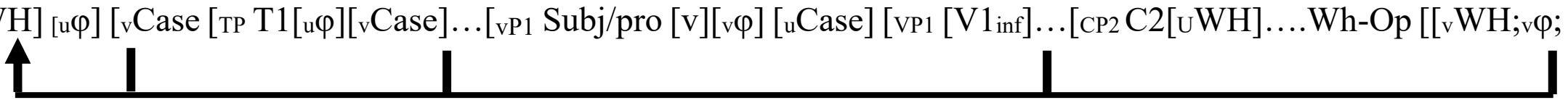

In a successive cyclic fashion, once a wh-question is formed, and once the topmost Wh-Op is (re)merged in the topmost Spec,CP, it forms an AM with all probes. As a result of this MATCHING mechanism, all C°s, (T°s) and *v*°s of lower phases get the same value, i.e., *v*F for all features including [WH] (as for Cs), provided that the syntactic environment is preserved.

With this in mind, we propose that YIA multiple wh-agreement structures like (13), i.e. consisting of at least 2 CP-phases, will have the underlying structure in (14).

(13) ʔayyan kitāb qāl ʕali ʔinn-uh xābar ʔabu-uh ʕann-uh?
Which book said.3MS Ali-NOM that-he told father-his about-it
‘Which book did Ali say that he told his father about?’

(14) [CP1 [ʔayyankitāb] C1 [TP1 [T1 qāl [vP1 ʕali [v [~~qāl~~] [CP2 ~~[ʔayyan kitāb]~~ [C ʔinn-uh [TP2 [T2 [xābar] [vP2 *pro* [VP2 [~~ʔayyan kitāb]~~ [V [~~xābar~~ ] [DP [ʔabu-uh] [~~PP~~ [P [ʕann] [DP ~~[ʔayyan kitāb]~~ ]]]]]]

Given (14), the derivation proceeds as follows. The Wh-Op *ʔayyan kitab,* as a d-linking wh-phrase[9], represented in our analysis as DP3, merges with the P *ʕann-uh* ‘about.it’ as a whole,

[9] D-linking, or discourse-linking, was introduced in Pesetsky (1987) to account for the unique syntactic behavior of specific wh-phrases like “which N” compared to non-D-linked phrases like “what.” Unlike “what,” “which N”

forming PP, which in turn merges with DP2 *ʔabu-uh* 'father-his' constituting V2 *xābar* 'told'. V2 merges with v2 forming Vp2, which merges with T2 constituting TP2. TP2 merges with C2 *ʔinn-uh* 'that-he' resulting in CP2. The same story iterates with regard to the derivation in CP1 layer. Given (4) and (6), C1 probes downwards and constitute MATCHING operation in which lower probes (T1), v1, C2, (T2), v2 get the vF values of the Wh-OP, namely wh-word/phrase (cf. Pesetsky and Torrego 2007). The agreement features on C1, are also spelled out on lower Cs, (Ts), vs, *pros*. Thus, AAP results in valuation of φ-features, which are spelled out as agreement inflection on verbs/Ts (*qāl*'s inflections here are null) (cf. also Alexiadou and Anagnostopoulou 2001; Boeckx 2003), and edge and copy features. Edge agreement features are spelled out as a result of Island effects on the possessive, or otherwise Clitic Construct State (CCS) *Abu-uh*, resulting in *-uh* attached to *Abu*. Given that CCS is a complex DP, any extraction from within it has to be "compensated for" (see e.g. Boeckx 2003, Bondevik et al. 2021; Chaves and Dery 2019; Pañeda et al. 2020; Abels 2019).[10]

As for edge features, we have here one of these, namely [WH], the valuation of which requires (re)emerging of an element like that of EPP (see e.g. Carstens 2005). Given that YIA is not a wh-in-situ language, C's [uWH] feature has to be valued, the Wh-Op *ʔayyan kitab* being in-situ, i.e. V-compl configuration, has to raise to Spec,CP2, and Spec,CP1. There will be a MATCHING relation between *ʔayyan kitāb*'s featured with all Cs' corresponding ones including [WH], by moving to Spec,C2, and Spec,C1 successively. This being the case, the goal *ʔayyan kitāb* (qua having [vWH], [uφ], *ʕali* (qua having [vφ], *pro* [vφ/vCase), when raising to their landing sites, will "activate" AAP, and MATCHING mechanism. As for the copy left after movement, given the copy theory, when the DP3 moves, it leaves a copy, and given that PP is an island, whose DP-internal position is a Case position, namely an Acc Case position, it then turns out that this copy is spelled out as a resumptive pronoun *–uh* (cf. Szabolcsi 2006; Szabolcsi and Lohndal 2017; Hein and Murphy 2020).

## 4. Grammaticalization of the suffixes *-uh, -eh, -um, -nen*

Attempting question (i) posed in section 1, namely what is the suffix *-eh/-uh*, *um*, or *-nen* on both the Wh-Op and verb?, understanding these suffixes resolves much of their ambiguity. Consider (15) from Classical/standard Arabic.

(15) a. man hua llað-i jaʔa ?
Who he that.3MS came.3MS
'Who came?'
b. man hia llati jaʔat?

---

phrases are constrained to a pre-established set of referents in the discourse, allowing them to bypass traditional syntactic islands that typically block wh-movement. This distinction significantly influenced the development of the minimalism, as it highlighted that syntactic movement is not solely governed by formal feature matching but is deeply sensitive to pragmatic and discourse-based information. Linguists have since utilized D-linking to explain variations in scope, island sensitivity, and the intervention effects that characterize complex sentence structures across diverse languages (see e.g. Ren and Yan 2024; Szabolcsi 2006).

[10] An alternative approach to certain apparent long-distance wh-dependencies treats them as instances of sluicing or related ellipsis phenomena rather than genuine successive-cyclic extraction (see e.g. Abels 2019; Agbayani and Ochi 2023; Alshaalan and Abels 2020). For example, Abels (2019) argues that apparent cases of "massive pied-piping" are not derived by exceptional wh-movement but instead result from recursive Contrastive Left Dislocation followed by clausal ellipsis. The YIA constructions examined here, however, do not lend themselves to such an analysis. The obligatory object clitic *-hen* together with the stranded floating quantifier *kull-hen* provide overt morphological evidence that the dependency is established within the VP domain before the wh-phrase reaches the left periphery. These diagnostics therefore favor a genuine movement analysis involving an intermediate VP-internal licensing position over an ellipsis-based derivation.

Who she that.3FS came.3FS
'Who came?'
c. man hum llað-iina jaʔuu ?
who they that-3MPL came.3MPL
'Who came?'
d. man hunna llāti jiʔnna?
Who they that.3FPL came.3FPL
'Who came?'

The pronouns *hua, hia, hum,* and *hunna* are considered comments/xabars whose subject/mubtadaʔ is *man* in (15a-d). Put differently, they are pronouns, i.e., lexical items. With this in mind, consider (16) from YIA.

(16) a. man-uh jaʔa ?
who-3MS came.3MS
'Who came?'
b. man-eh jaʔa*(h)?
Who-3FS came.3FS
'Who came?'
c. man-um jaʔu *(um) ?
Who-3MPL came.3MPL
'Who came?'
d. man-nen jaʔ*(ayn) ?
Who-3FPL came.3FPL
'Who came?'

In YIA data presented in (16), *–uh, -eh, -um,* and *-nen* seem to be greatly different than similar to those presented in (15), namely *hua, hia, hum,* and *hunna*. They are different in terms of: i) they are not "free" pronouns. This is clear from the fact that the Wh-Op *man* + one of them together function as the subject/topic of the verb *jaʔa*, ii) no relative pronoun is needed. In (15), relative pronouns such as *llaði* 'that/who', is an obligatory element, which is not the case in (16), iii) the 'h' sound is an obligatory element in (15), but optional in (16),[11] and iv), *–u(h), -e(h), -um,* and *-nen* appear on both the Wh-Op and the verb, which is not the case with *hua, hia, hum,* and *hunna*. And this appearance is not haphazard, but rather systematic, as if one appears only on one element, viz., the Wh-Op or the verb, the sentence will be rendered ungrammatical. From all (i-iv), it follows that *–uh, -eh, -um,* and *-nen* are not pronouns in this case, but rather agreement inflection markers.

It also follows that the SA/CLA pronominal forms have certainly undergone a grammaticalization process. Grammaticalization process is a diachronic linguistic process where a lexical element becomes a grammatical one (see e.g. Lucas 2015; Hansen and Visconti 2014). These elements are thus not resumptive pronouns in the syntactic technical sense. To differentiate them from resumptive pronouns, consider (17).

(17) a. kull-u ar-rijāl-i jaʔuu.
all-NOM the-men-GEN came
'All the men have come.'
b. ar-rijāl-u kull-hum jaʔuu
the-men-NOM all-NOM came
'All the men have come.'

[11] Note that while the 'h' sound is a word initial element in (15), it is word-final in (16).

In (17a), the DP *kull-u ar-rijāl-i* is a CS whereby *kullu* is the head N and *ar-rijāli* is its complement. It functions as the subject of the verb *jaʔuu*. However, when *ar-rijāli* undergoes a head to Spec movement, and since a CS is an island, i.e., no movement is possible, the resumptive pronoun -hum is spelled-out so that no island principle is violated. Note also that -hum agrees with *ar-rijāl* in all φ-features. Although this sort of agreement is DP-internal, i.e. between elements of the same DP, this is somehow similar to the agreement in (16) where the subject agrees with the verb, i.e. DP-external, marked with *-uh*, *-eh*, *-um*, and *-nen*, respectively. Further evidence can be provided from complex wh-questions in YIA as in (18).

(18) man-eh qāl ʕali ʔinn-uh qābal-eh?
who-3FS said.3MS Ali that-3MS met-3FS
'Who did Ali say that he met?'

In (18), we have strong evidence in support of our argument. The Wh-Op *man-eh* agrees with the embedded verb *qābal-eh*. In the matrix clause, however, there is another type of agreement, which is manifested among the matrix verb *qāl,* the subject *ʕali* and the complementizer *ʔinn-uh.*

We can conclude that the *–uh, -eh, -um,* and *-nen* are inflectional agreement markers/ that mark agreement between the Wh-Op and the verb (and subject). However, the verb *qābal-eh* in (18) is transitive; the *-eh* here marks accusative Case. In this sense, these elements behave like resumptive pronouns.

## 5. Wh-agreement in YIA: An AAP approach

We begin this section by noting that unlike SA, where the subject agrees with the verb in person and gender features in VSO, in YIA the subject agrees with the verb in all φ-features in both SVO and VSO as (19) illustrates regardless of the word order like almost all modern dialects of Arabic (see e.g. Aoun et al. 2010).

(19) a. jāʔ-u l-ʕyiāl
came-3MPL the-boys
'The boys came.'
b. l-ʕyiāl jāʔ-u.
the-boys came-3MPL
'The boys came.'

The same agreement facts are also manifested in wh-questions, where the subject wh-word agrees with the verb. Not only does subject wh-word agree with the verb in YIA, but also object wh-words. To repeat a point stated earlier, wh-words in YIA are *man* "who", *mu* 'what', *kayf* 'how' and *ʔayn* 'where'. These words are roots/bases to which the agreement suffixes, viz. -*uh*, *-eh, -um, -(h)en*, etc. are attached.

### 5.1. Wh-agreement in one-phase questions

The data presented below illustrate this phenomenon in which the wh-word morphologically agrees with the verb in all φ-features. This agreement is morphologically marked by overt or covert morphemes on C, T and V. The data are presented in (20) below (which considers "who-subject").

(20) a. man-u(h) jāʔ ?
who-3MS came.3MS
'Who came?'
b. man-eh jāʔ-ah?

who-3FS came-3FS
'Who came?'

c. man-en jāʔ-ayn ?
who-3FPL came-3FPL
'Who came?'

d. man-um jāʔ-um ?
who-3MPL came-3MPL
'Who came?'

[CP man-um [C ] [TP [T jāʔ-um] [ vP [~~man-um~~ [ v ~~jāʔ-um~~] [ VP [V ~~jāʔ-um~~] ]]]]

The data in (20) present wh-agreement of the wh-word *man* 'who' as a subject. As can be observed in (20a – 20c), there is an agreement between the wh-word and the verb in all φ-features, (i.e. number, gender and person). In (20a), for example, *man-u(h)* is singular, masculine and 3rd person and the verb *jāʔ* 'came' is singular, masculine and 3rd person, hence (20a) is grammatical. The same, albeit different, is true for (20b) in which *man-eh* is singular, feminine and 3rd person and the verb *jāʔ-ah* 'came' is singular, feminine and 3rd person, hence (20b) is grammatical. The same can be said about (20c & d), wh-word *man* 'who' and the verb agree in all φ-features. Now, consider examples in (21) when wh-word *man* 'who' is an object.

(21) a. man-u(h) (li)qabal-uh ʕali ?
who-3MS (that) meet-3MS Ali
'Who did Ali meet?'

b. man-eh (li)qabal-eh ʕali?
who-3FS (that) meet-Acc-3FS Ali
'Who did Ali meet?'

c. man-eh (li)qabal-ah ʕali?
who-3FS (that)meet-Nom-3FS Ali
'Who did meet Ali?'

[CP man-eh [C ] [TP [T qabal-ah] [ vP [ v ~~qabal-ah~~] [ VP [V ~~qabal-ah~~] [DP ʕali]]]]]]

In (21a) and (21b), "who" is an object, but it is a subject in (21c). On the one hand, the wh-word *man-uh* 'who-M' in (21a) first merges in V-compl configuration, precisely DP and then moves to Spec,VP then to Spec,CP. The same holds true for the wh-word *man-eh* 'who-F' in (21b), i.e. it merges in DP and then moves to Spec,VP then moves to Spec CP. As for agreement, the verb *liqabal-uh* 'meet' in (21a) agrees with the wh-word *man-uh* in all φ-features, viz., 3rd person, masculine and singular. These agreement features are all represented in the inflection *-hu* in both constituents. Further, the suffix *–uh* does not only show an agreement between the wh-word and the verb but also it shows a structural accusative case.

The same thing can be said about the agreement phenomenon in (21b), i.e. the verb *qābal* has the agreement inflection*–eh* which shows agreement features, namely 3rd person, feminine and singular between the wh-word *man-eh* and the verb *liqābal-eh* as well as accusative case. On the other hand, though these φ-features remain the same between the wh-word and the verb, the agreement inflection *–ah* on the verb *qabal,* shows, in addition to agreement, a nominative case. This difference in case marking can be noticed by looking at the difference between *-eh* and *-ah*, where *e* indicates accusative and *a* nominative.

The question here is: which element/feature licenses Acc/Nom case in YIA? Does *li-* has some features that assign Acc case/Nom case? Chomsky (2005, 2006) argues that the locus of Nom Case and agreement is $C^0$. If so, why the verb *qabal-ah* assigns Nom case. According to Al-Balushi (2011: 190), both Nom and Acc Cases might actually be licensed as a result of $T^0$ and $v^{*0}$, respectively, entering an Agree relation with $Fin^0$. Put simply, $Fin^0$ is the locus of both Nom and Acc Cases, thus licensing in general. Consider the examples given in (22), which address what, specifically those involving the relativize *li-*[12]

(22) a. mu-hu (li-*)ʔaštri-uh ʕali ?
what-3MS (that) bought-it-3MS Ali
'What did Ali buy?'
b. mu-he(h) (li-*)ʔaštri-eh ʕali ?
what-3FS (that) bought-it-3FS Ali
'What did Ali buy?'
c. mu-m (li-*)ʔaštri-hum ʕali ?
what-3MPL (that) bought-them-3MPL Ali
'What did Ali buy?'
d. mu-en (li-*)ʔaštri-hen ʕali ?
what-3FPL (that) bought-them-3FPL Ali
'What did Ali buy?'

The data in (22) present wh-agreement of the wh- word *mu* 'what' as an object. As can be observed in (22a) – (22d), there is an agreement between the wh-word and the verb in all φ-features, i.e. (number, gender, and person). Agreement goes fine and has no violations, but complexity lies in where and when agreement takes place.

---

[12] Aoun and Choueiri (1997) and Choueiri (2002) claim that relative complementizer enter into indefiniteness agreement relation with the relativized category. The relative complementizer is exclude when the relativized category is indefinite. What they conclude is that the complementizer *yalli* occurs only to relativize a definite noun. Aoun (2000) argues that *yalli* 'that' in Lebanese Arabic is a complementizer which bears the features {[+definite], φ-features, case}. Consider the following examples:
(i) ʔāl kariim ʔənna/*ʔənno bʔl-beet
said.3SM that.3SF in-the-house
'Karim said that she is in the house.'
(ii) l-bənt yalli bʔəl- beet
the-girl that in-the-house
'The girl that is in the house.'
Aoun indicates that *yalli* in (ii), like *ʔənno* in the above example, bears the necessary φ-features. He argues that (i) is well-formed because *ʔənno* bears φ-features that identify the embedded null subject. Brustad (2000) uses individuation hierarchy to solve the problem of using *illi* with indefinite noun. She argues that when nouns are more individuated the more it is likely to induce definite *illi*.

As for *li-* in YIA, we agree with (see Galal 2004) that it behaves like a relativizer, perhaps indicating an aspectual or temporal anchor within the clausal structure. Its behavior being obligatory with *mu-hu* (what) but optional with *man-uh* (who) suggests that it serves as a "tense-aspect facilitator" or a "verbalizer marker" that bridges the gap between the *wh*-operator in the Specifier position of the CP and the TP (Tense Phrase). It has the following syntactic properties: i) C-element. It behaves like a complementizer-like element: *li-* acts as a bridge between the wh-operator and the tense structure of the clause, ii) there is some sort of aspectuality/temporality. If our argument regarding FI is on the right track, then it seems that li- could function as aspectuality/temporality anchor, and iii) obligatoriness/optionality.

Given that *li-* is optional with the wh-Op *man-uh*, but obligatory with *mu-hu*, it seems that it has 3 features, viz., [+/- Human], [+/- animate] and [+/-specificity]. The latter is in line with (22) who uses the term "individuation" which, in other words, signals specificity. However, given that *mu-hu* is [-Human] and often targets an event or an object, which is generally generic, we propose that li- is needed to individualize/spiffy the entity to which the wh-Op refers to. From discourse perspective, given that *man-uh* is inherently [+human/+specific], it may interact with the verb's features directly, making the presence of *li-* redundant but permissible. However, *mu-hu* is -Human/-specific and -animate. Similarly, inanimate *wh*-elements require a "bridge" to the verbal event to ensure the clause is properly formed, which is why *li-* becomes mandatory. Another property distinguishing li- from its sisters in EA, LA, MA is that *li-* in YIA is behaves like a clitic cliticizing to the verb. Thus, if our argument is on the right track, we propose that *li-* is a C-element; when it is present it fills C, and the verb raises from V to v, Fin, and finally to C which it attaches to *li-*, and when it is absent/optional, C is [Φ].

In YIA, even *kayf and ʔayn*[13] ‘how and where’, respectively, manifest wh-agreement with the verbs. The data in (23 & 24) show that the *kayf/ʔayn* 'how/where' fully agree in all φ-features ((number, gender, and person). Any violation to any of these features leads to ungrammatically.

(23) a. kayf-u(h) ʕali ?
How-3MS Ali
How is Ali?’
b. kayf-eh ʕalia ?
how-3FS Aliya
‘How is Aliya?’
c. kayf-um l-ʕyāl ?
how-3MPL the-boys
‘How are the boys?’
d. kayf-en l-banāt ?
how-3FPL the-girls
‘How are the girls?’
e. kam-en li-jaʔayn ?
How-many-3FPL that-came.3FP
‘How many (girls) came?’
f. kam-um li-jaʔauu ?
How.many-3MPL that-came.3MPL
‘How many (boys) came?’

(24) a. ʔayn-u ʕali ?
Where-3MS Ali
‘Where is Ali?’
b. ʔayn-eh ʕalia ?
where-3FS Aliya
‘Where is Aliya?’
c. ʔayn-um l-ʕyāl ?
where-3MPL the-boy
‘Where are the boys?’
d. ʔayn-en l-banāt ?
where-3FPL the-girls
‘Where are the girls?’

However, if one of these agreement features is missed, the sentence is rendered ungrammatical. Consider (25).

(25) a. * man-u qābal-eh ʕali ?
who-3MS meet-3FS Ali
b. * man-um liqābal-nen ʕali ?
who-3MPL meet-3FPL Ali
c. * man-nen liqābal-um ʕali ?
who-3FPL meet-3MPL Ali

In (25a), for example, *man-u* is 3rd person, singular, and masculine, while *-eh* is 3rd person singular, and feminine, hence (25a) is ungrammatical. In (25b), *man-um* is 3rd person, plural, and masculine whereas the *-nen* in the verb is 3rd person, plural, and masculine, hence (25b) is ungrammatical due to the violation of one feature (i.e., number). The same thing can be said about (25c). Thus, we conclude that these agreement inflections are very crucial in YIA and any violation to one of the agreement features renders the structure ungrammatical.

[13] The only wh-word that does not show agreement is *mata* ‘when’.

### 5.2. Wh-agreement in multiple questions/wh-movement

In (15) through (25), we presented and discussed wh-agreement in simple wh-structures with *who, what, how*, when and *where* in subject, object position. However, YIA wh-agreement involves complex structures where there are multiple phase layers, viz., a wh-agreement in which the wh-word agrees with Cn and Tn, vn, or otherwise, there are multiple wh-movements. We first start with subject Wh-Ops in 2 CPs. Consider (7) repeated here as (26) for convenience.

(26) a. man-u qāl ʕali ʔinn-uh jāʔ ?
who-3MS said.3MS Ali that-3MS came-3MS
'*Who did Ali say that he came?'
b. man-eh qāl ʕali ʔinn-eh jāʔ-ah ?
who-3FS said.3MS Ali that-3FS came-3FS
'*Who did Ali say that she came?'
c. man-en qāl-ah ʕalia ʔinn-en jāʔ-ayn ?
who-3FP said-3FS Aliya that-3FPL came-3FPL
'*Who did Alia say that they came?'

[CP1 man-en [C ] [TP1 [T qāl-ah] [ vP1 ʕalia [ v ~~qāl-ah~~] [ VP1 [V ~~qāl-ah~~] [CP2 [~~man-en~~] [C ʔinn-en] [TP2 [T jāʔ-ayn] [ vP2 [~~man-en~~] [ v ~~jāʔ-ayn~~] [VP2 [V ~~[jāʔ-ayn~~]]]]]]]]

In (26a-c), for example, the wh-words (or C1) *man-u, man-eh, man-en* agree with C2, T2 and v2 *ʔinn-uh, ʔinn-eh, ʔinn-en* and T2/V2 *jāʔ, jāʔa-ah, jāʔ-ayn*, *pro* respectively. However, T1/V1 does not agree with them; rather it agrees with the DP-subj *ʕali/ ʕalia*. The fact that C1 agrees with C2, T2, v2, while T1/V1 agrees with each other is a solid consequence of (6') and (11). Empirically, this is straightforwardly clear if we consider the valuer of C1 and C2. Simply put, given that the Wh-Op *man-en* is the same valuing element of both phase heads, C1 and C2, both heads have to carry the same features. This is also clear from the ungrammaticality of structures like (27a) in which the verb *qāl-ah* (3FS) does not agree with its subject *ʕali* (3MS). The same holds true also for any violation to any clause in (11), as is in (27b), where *C1* (3FS) does not agree with C2 (3MS).

(27) a. *man-u qāl-ah ʕali ʔinn-uh jāʔ-a ?
who-3MS said-3FS Ali that-3MS came-3MS
b. *man-eh qāl ʕali ʔinn-uh jāʔ-a ?
who-3MS said.3MS Ali that-3MS came-3MS

Now, we turn to wh-agreement in object constructions in more than 2 CPs, where a wh-question consists of multiple wh-movement as in (28).

(28) a. mu-uh li-qāl ʕali ʔinn-uh ʔaštri-uh ?
what that-said Ali that-he bought-3MS
'*What did Ali say that he bought it?'
b. mu-hen li-qāl ʕali ʔinn-uh ʔaštri-hen ?
what that-said Ali that-he bought-them
'*What did Ali say that he bought them?'

[CP1 mu-hen [C ] [TP1 [T liqāl [ vP1 ʕali [ v1 ~~liqāl~~] [ VP1 [~~mu-h~~en] [V ~~liqāl~~] [CP2[~~mu-h~~en] [C ʔinn-hu] [TP2 [T ʔaštri-hen] [ vP2 pro [v ~~ʔaštri-hen~~] [VP2 [~~mu-h~~en] [V ~~ʔaštri-hen~~] [DP [~~mu-h~~en]]]]]]]]]]

In (28a), the object wh-word moves from V-compl position to Spec,VP2, Spec,CP2, and then to Spec,CP1. Examples (28) involve complex structures where there are multiple phase layers, viz., wh-agreement in which the wh-word agrees with $C_n$ and ($T_n$), $v_n$. Given (6’ & 11), the topmost C agrees with $C_n$, ($T_n$) and $v_n$, provided that the clause AM is applied to phase heads, while FV is applied to nonphase heads, depending on the syntactic environment.[14]

There is also some kind of UG variation in wh-questions, specifically those containing more than 1 CP phase. Consider (29), for instance, from YIA and their English glosses (29).

(29) a. ʔayyan rajjāl qāl ʕali *(ʔinn-uh) qābal-uh ʔams ?
which man said Ali that-he came yesterday
‘Which man did Ali say (*that he) came yesterday?’
b. ʔayyan kutub qāl ʕali *(ʔinn-hen) bi-l-maxzan?
which books said Ali that-they in-the-store
‘Which books did Ali say (*that they) in the store?

In (29), it is clear that while “that-RP” in YIA is obligatory, in English it is not permitted. This could lead to the optionality of CP2 in English multiple-phase wh- questions. However, given that Spec,vP is not an A’-argument position (qua our synthesis), a CP2 null is a must.

The final case which we would like to address in this section is an example of multiple wh-questions and coordination as in (30).

(30) ʔayyan kitāb qāl ʕali ʔinn-uh ʔaddy-uh liman dār ħuwār bayn-uh
Which book said-3MS Ali-NOM that-3MS gave-he to.whom had a talk between-him
wa bayn moħammad ʕann-uh?
and between Mohammed about it?
‘Which book did Ali say that he gave to who had a ward with Mohammed about?’

[CP1 [ʔayyan kitāb] C1 [TP1 [T1 [qāl] [vP1 ʕali] [v ~~[qāl~~] … [CP2 [~~[[ʔayyan kitāb]~~] C [ʔinn-uh [ [TP2[T2 [ʔaddy-uh[vP2 pro] [v [VP2 ~~[ʔayyan kitāb]~~ [V[~~ʔaddy-uh~~] [pp [p li- [CP3 man [C3 [TP3 [T3 dār [vP3 ħuwār [&P [pp [P bayn-uh [wa [PP [P bayn [DP [N moħammad [~~PP~~ [P [ʕann] [DP ~~[ʔayyan kitāb]~~ ]]]]]]

(30) is perhaps the longest wh-question in this study. It is, in other words, an example of how (4, 6’, 11) together work. More importantly, it is indicative of how human $C_{HL}$ works in PT framework. Given (11) and (6’), there are four major steps we will focus on: 1) AM, 2) FV, and 3) Islands and Copy, 4) Case. As for AM C1 will match (T1), (pro1), v1, C2, (T2), (pro2), v2, C3, (T3), (pro3), and v3. Regarding FV, it will work only for nonphase heads in some syntactic environments. There are 3 such environments in (30), namely [qāl ʕali], [dār ħuwār], and [bayn-uh]. *ʕali* raises from DP1 (within the &P), to Spec,vP2, given that Spec,vP3 is occupied by *ħuwār*. During its raising journey, *ʕali* stops in Spec,vP2, and leaves *pro*. It also passes by the complement of the P *bayn*, as one element in the coordination structure *bayn ʕali wa muhammad* ‘between Ali and Mohammed’ and since PP is an island (step 3), its movement results in RRP *-uh* as a result of (joint working mechanism between) FV, island (step 3), and

[14] Note that YIA allows ʔayyan ‘which’ to appear alone (functioning as a topic) in wh-questions. This is exemplified in (i):

(i) ʔayyan-u li-jāʔ ?
Which-3MS that-came.3MS?
‘Who came?’

(ii) ʔayyan-nen li-jāʔayn ?
Which-3FPL that-came.3FPL?
‘Who came?’

And when it occurs in this use, it carries agreement in all φ-feature, exactly like *man* ‘who’, for example.

Case (step 4). Given AM, the wh-phrase *ʔayyan kitab* first merges as the object of the P "about", it raises to Spec,VP2. When it raises from there, it leaves a copy, which gets spelled out as *-uh* attached to the verb *ʔaddy*. From there it raises to Spec,CP2, and since Spec,CP2 is a A'-argument (has no Case), the copy is not spelled out. It finally raises to Spec,CP1, its final landing site.

**6. AAP and wh-agreement cross-linguistically**

In this section, we apply our proposal to wh-movement across languages. As we have discussed in previous sections, wh-agreement has been referred to as an agreement morphologically marked on either the verbs or complementizers or both, and is caused by wh-movement (see e.g. Lochbihler and Mathieu 2016; Jeoung 2020, and references therein). Specifically, we apply AAP approach to Greek, Indonesian, Palauan and Irish. To start with, consider (31) from Greek.

(31) a. pia enai i yanika eki;
Who-3FS is the woman-3FS there
'Who is the woman there?'
b. pios enai o anthrwpos eki;
who.3MS is the man there
'Who is the man there?'
c. pion gnórise i Phoevos;
Who.ACC.3MS met.3MS the Phoevos
'Who did Phoevos meet?'
d. pios synántisa oi gynaíkes;[15]
Who.ACC.3MS met.3MS the women
'Who did the women meet?'

[CP pios [C ] [TP [T synántisa] [ vP [ v ~~synántisa~~] [VP [V ~~synántisa~~] [DP oi gynaíkes]]]]]]

Examples in (31a-d) involve only 1 CP phase where AAP takes place between C, T, v. In (31a), the Wh-Op *pia* 'who' agrees with the verb *enai* 'is' in all φ-features: both are 3rd person feminine singular. However, in (33b) both are 3rd person masculine singular. In (33c & d), the wh-words *pion* and *pios* 'whom' function as the objects of the verbs *gnórise* and *synántisa* 'met', respectively, as they are transitive. The wh-agreement in (31a-d) takes place within one phase, viz., CP, given (4) and (6). The first pair (31a & b) represents subject wh-movement, i.e. from Spec,vP to Spec.CP1. Given that the Wh-Op in (31c & d) functions as the object of the verb, the wh-movement takes place from V-Compl position to Spec,VP, and finally Spec,CP.

In Greek, specifically modern Greek, the Wh-Op is also marked for Case. Consider (32), which represents simplex wh-questions, i.e. containing 2 CPs.

(32) a. Pji ipe oti efighan?
Who.NOM-PL said.3SG that left.3PL
'*Who did he say that left?' (cf. Who did he say left?)
b. Pjion ipes oti idhes?
Who.ACC-SG said.2SG that saw.2SG
'Who did you say (that) you saw?'
(Tsimpli and Dimitrakopoulou 2007: 220).

[15] Note that in Greek the ';', at the end of question is a question mark, the same like '?'.

In (32a) the verb *efighan* 'left' agrees with the wh-word *pji* 'who' in all φ-features. The wh-word *pji* 'who' and the verb *efighan* 'left' are in 3rd person plural forms. The verb *ipe* 'said', which is in 3rd person singular, agrees with the null/*pro* subject which is part of the bound embedded clause. The Wh-Op *ipe* also carries case, i.e. nominative (Timpli and Dimitrakopoulou, 2007). Like Arabic, Greek is a null subject language. However, in (32b) wh-word *pjion* 'whom' carries an accusative case, but it may not agree with it in φ-features, at least one feature, viz., person, unlike YIA.

A short note about Greek wh-agreement phenomenon is in order here. In this language, resumptive pronouns and complementizers behave differently, say, from YIA. Consider (33).

(33) a. Pjon ipes oti (ton) prosevalan xoris logho?
whom said.2SG that him-insulted.3PL without reason
'Who did you say that they insulted (*him) without a reason?'
b. Pjon fititi ipes oti (ton) aperipsan sti sinedefksi?
which student.3MS said.2SG that him-rejected.3PL at-the interview
'Which student did you say that they rejected at the interview?'
(Tsimpli and Dimitrakopoulou, 2007: 220).

In these examples, the resumptive pronoun (e.g. *ton* 'him') is optional in Greek, as (33a & b) demonstrate. The complementizer *oti* 'that' is not optional in both examples. Additionally, AAP accounts for these factors straightforwardly. Put differently, given that both (33a & b) contain multiple phases, precisely C1, C2, (T1), (T2), v1, v2, which form a matching mechanism under AAP. When the topmost probe, i.e. C1 probes for the DP in its search space, namely *Pjon fitit i* an AAP operation is established between them. As soon as C1's features get valued, all other probes' features get valued, interpreted (and eliminated). Thus, we can conclude that AAP fits well to account for wh-agreement phenomenon in Greek (see also Panagiotidis 2010).

Now, consider examples (34) from Indonesian, Palauan, and Irish, respectively.

(34) a. Adik mem-baca buku siapa? Indonesian
sibling ACTV-read book who
'Whose book is brother reading?'
b. ng-te'ai [a kileld-ii a sub ei] Palauan
CL-who R-PF-heat-3SG soup
'Who heated up the soup?'
c. Cé aL chonaic tú? Irish
who COMP.L saw you
'Who did you see?'

In (34a), the wh-expression *siapa* 'who' appears as the possessor of *buku* 'book'. Although Indonesian permits wh-in-situ questions, the interrogative dependency is nevertheless established syntactically. Under the AAP approach, the derivation contains the phase heads C and v, which form a MATCHING chain. Once the topmost C searches its c-command domain and matches the wh-expression *siapa*, an AAP relation is established between them. It follows that Cs unvalued [WH] and φ-features are valued, and the same valued features are transmitted cyclically to the remaining phase heads through Agree as MATCHING mechanism. Thus, even in the absence of overt wh-movement, the interrogative dependency is licensed through successive valuation of phase heads. It follows that AAP straightforwardly accounts for the underlying syntactic dependency in Indonesian, while the special morphology found in extraction contexts can be viewed as the morphological realization of this agreement relation (see also Abels and Dayal 2023).[16]

As for (34b), from Palauan, the wh-operator *te'ai* 'who' occupies Spec,CP, while the verb appears with the characteristic morphology associated with subject extraction. Under the AAP analysis, the derivation consists of the phase heads C and v, which are permanently linked through successive Agree relations. Once the highest probe C establishes a MATCHING relation with the Wh-Op in Spec,CP, its unvalued [WH] and φ-features get valued. These valued features are then inherited by all lower phase heads, including v, thereby creating a uniform agreement chain across the clause. Consequently, the special verbal morphology associated with subject extraction is interpreted as the overt reflex of this cyclic valuation process rather than as an isolated movement-triggered phenomenon. Hence, AAP provides a principled account of Palauan wh-agreement by deriving the observed morphology from feature valuation across phases.

The AAP approach seems also to account for Irish data. In (34c), the wh-expression *Cé* 'who' occupies Spec,CP, while the complementizer is realized as *aL*, triggering lenition on the following verb. Under AAP, the derivation contains the phase heads C, T and v, which participate in a permanent agreement chain. Once the highest probe C establishes an Agree MATCHING relation with the moved wh-expression, its [WH] and φ-features are valued. The valued features are subsequently transmitted to all lower phase heads, yielding a single agreement chain throughout the derivation. The complementizer alternation observed in Irish is therefore analyzed as the morphological spell-out of the valued features on C, a result of wh-movement. Thus, AAP captures the Irish pattern by attributing the complementizer alternation to AM across phase heads.

To recapitulate, when comparing YIA to languages like Greek, Indonesian, Palauan, etc., it becomes evident that YIA exhibits a more robust and pervasive system of wh-agreement. While these cross-linguistic counterparts typically manifest agreement through limited morphological markers on the verb or complementizer, or both, or via optional resumptive pronouns, YIA displays a systematic AAP mechanism that triggers morphophonological marking across the entire functional hierarchy starting from the Wh-Op itself walking through Cs, (Ts) and vs. In YIA, the appearance of the resumptive pronoun *-uh* (or its sisters) is not merely an optional strategy for island-crossing, as seen in Greek, for instance, but a mandatory consequence of long-distance agreement that ensures the survival of the derivation at both PF and LF. Unlike the singular or partial agreement patterns often observed in Indonesian or Irish, the YIA dialect demonstrates a dense, multi-layered agreement chain where φ-features are inherited and spelled out in every position of the wh-dependency. Thus, YIA provides stronger empirical support for the AAP approach, as it reveals a higher degree of morphological integration between the moved operator and the functional heads in the derivation, positioning it as a primary case study for the richness of wh-agreement in Arabic vernaculars.

[16] Jeoung (2020: 121) provides empirical evidence from Indonesian that even in possessive DPs, wh-agreement can be marked on the moved possessor. The extracted possessive DP can undergo A'-movement like other nominal arguments. These are shown in (i) and (ii).

(i) Rumah Lani di-rata-kan kemarin.
house Lani PASS-flat-APPL yesterday
'Lani's house was destroyed yesterday.'

(ii) Siapa yang rumah-nya/*rumah __di-rata-kan kemarin?
who COMP.FOC house-DEF PASS-flat-APPL yesterday
'Whose house was destroyed yesterday?'

Jeoung considers these possessive DP phases having a escape hatch, edge. Like PCs,vPs, and when the extracted element undergoes an A'-movement, wh-agreement is morphologically marked.

## 7. Conclusions and implications

This study has provided a formal account of wh-agreement in YIA, establishing that the dialect employs a sophisticated system of morpho(phono)logical marking across functional heads. In the course of this study, we have proposed auxiliary mechanisms: i) reconsidering locality conditions and relativized minimality constraints in minimalist syntax, we propose that the Wh-Op has to move to Spec,VP prior to targeting Spec,CP, a requirement driven by several factors including Case-licensing, Acc Case and locality constraints, ii) looking at the nature of RRP *li-*, we propose that in YIA it acts as a relativizer clitic, considering it a C-element. It serves as an aspectual/temporal anchor. Unlike its counterparts in other Arabic dialects, *alli, illi,* etc. Its obligatoriness with inanimate *mu-hu* ([-Human/-Specific]) vs its optionality with human *man-uh* ([+Human/+Specific]) suggests that *li-* functions to individuate and specify entities, bridging the gap between the Wh-Op and the TP. We propose that when *li-* is present, it occupies C, attracting the verb through V-to-v-to(-Fin-to)-C movement, whereas its absence implies a null C head,[17] and iii) given that $C_{HL}$ utilizes a multi-step computation, viz., multiple wh-movements, to satisfy both morphological and syntactic requirements before reaching the topmost CP, we suggest that PT could be part of I-language architecture. This structure allows us to observe the mechanics of the $C_{HL}$ with greater clarity.

Equally important is the study contribution to the debate surrounding UG by presenting novel empirical evidence from YIA wh-agreement patterns. These patterns serve as a critical test case for the existence of innate syntactic constraints, demonstrating that such agreement operates independently of input frequency. By revealing how specific I-language operations reflect deeper, invariant principles of human cognition, our findings reinforce the UG model, validating the hypothesis that human language is structured by a universal, biologically determined framework rather than being solely derived from environment-specific learning (see also Chomsky 2021).

Our study vividly demarcates that investigating under-investigated dialects act as critical "microcosms" that provide more subtle, robust evidence for universal syntactic principles than well-studied languages. These findings carry significant implications for the future of NLP. As the field moves toward more generalized architectures, integrating the nuanced syntactic "secrets" of under-resourced languages like YIA is essential. By incorporating these rich, non-standard datasets, we can train NLP models including MARBERT, XLM-R that on low-resource dialects like YIA to finetune them to align with the underlying, biologically determined principles of human language architecture.

[17] Fin is proposed in case we project the CP layer into TopP, FocP, ForcP, and FinP (cf. Rizzi 1997, 2004).